\documentclass[conference]{IEEEtran}
\IEEEoverridecommandlockouts

\usepackage{cite}
\usepackage{amsmath,amssymb,amsfonts}
\usepackage{algorithmic}
\usepackage{graphicx}
\usepackage{textcomp}
\usepackage{xcolor}
\usepackage{url}
\usepackage{comment}
\def\BibTeX{{\rm B\kern-.05em{\sc i\kern-.025em b}\kern-.08em
    T\kern-.1667em\lower.7ex\hbox{E}\kern-.125emX}}

\begin{document}

\title{Optimizing Medical Question-Answering Systems: A Comparative Study of Fine-Tuned and Zero-Shot Large Language Models with RAG Framework\\

}


\author{
    \IEEEauthorblockN{Tasnimul Hassan}
    \IEEEauthorblockA{
    \textit{Department of Electrical Engineering}\\ \textit{and Computer Science} \\
    \textit{University of Toledo} \\
    Toledo, USA \\
    tasnimul.hasan@rockets.utoledo.edu}
    
    \and
    
    \IEEEauthorblockN{Md Faisal Karim}
    \IEEEauthorblockA{
    \textit{Department of Electrical Engineering}\\ \textit{and Computer Science} \\
    \textit{University of Toledo} \\
    Toledo, USA \\
    mdfaisal.karim@rockets.utoledo.edu}
    
    \and
    
    \IEEEauthorblockN{Haziq Jeelani}
    \IEEEauthorblockA{
    \textit{Institute of Mathematical Sciences} \\
    \textit{Claremont Graduate University} \\
    Claremont, USA \\
    haziq.jeelani@cgu.edu}
    
    \and
    
    \IEEEauthorblockN{Elham Behnam}
    \IEEEauthorblockA{
    \textit{Department of Bioengineering} \\
    \textit{University of Toledo} \\
    Toledo, USA \\
    elham.behnam@rockets.utoledo.edu}
    
    \and
    
    \IEEEauthorblockN{Robert Green}
    \IEEEauthorblockA{
    \textit{Department of Computer Science} \\
    \textit{Bowling Green State University} \\
    Bowling Green, USA \\
    greenr@bgsu.edu}
    
    \and
    
    \IEEEauthorblockN{Fayeq Jeelani Syed}
    \IEEEauthorblockA{
    \textit{Department of Electrical Engineering}\\ \textit{and Computer Science} \\
    \textit{University of Toledo} \\
    Toledo, USA \\
    syedfayeq.jeelani@rockets.utoledo.edu}
    }
\maketitle

\begin{abstract}
Medical question-answering (QA) systems can benefit from advances in large language models (LLMs), but directly applying LLMs to the clinical domain poses challenges such as maintaining factual accuracy and avoiding hallucinations. In this paper, we present a retrieval-augmented generation (RAG) based medical QA system that combines domain-specific knowledge retrieval with open-source LLMs to answer medical questions. We fine-tune two state-of-the-art open LLMs (LLaMA~2 and Falcon) using Low-Rank Adaptation (LoRA) for efficient domain specialization. The system retrieves relevant medical literature to ground the LLM's answers, thereby improving factual correctness and reducing hallucinations. We evaluate the approach on benchmark datasets (PubMedQA and MedMCQA) and show that retrieval augmentation yields measurable improvements in answer accuracy compared to using LLMs alone. Our fine-tuned LLaMA~2 model achieves 71.8\% accuracy on PubMedQA, substantially improving over the 55.4\% zero-shot baseline, while maintaining transparency by providing source references. We also detail the system design and fine-tuning methodology, demonstrating that grounding answers in retrieved evidence reduces unsupported content by approximately 60\%. These results highlight the potential of RAG-augmented open-source LLMs for reliable biomedical QA, pointing toward practical clinical informatics applications.
\end{abstract}


\section{Introduction}
Large language models (LLMs) have dramatically advanced the state of natural language understanding and generation. The release of GPT-3 \cite{brown2020gpt3} demonstrated that very large LLMs can achieve remarkable few-shot question-answering performance. More recently, open-source LLMs such as LLaMA~2 \cite{touvron2023llama} and Falcon \cite{almazrouei2023falcon} have shown strong capabilities, though they typically lag behind the latest proprietary models in raw performance. This trade-off is often acceptable given their accessibility and customizability for domain-specific applications. There is particularly strong interest in applying LLMs to the biomedical and health informatics domains, where these models could assist clinicians and patients in answering questions by leveraging the vast body of medical knowledge \cite{zhao2023survey,zhou2024survey}. Early explorations have shown promising results; for example, Google's Med-PaLM achieved 67.2\% on USMLE-style questions, becoming the first to exceed a passing score \cite{singhal2023medpalm}, while GPT-4 has demonstrated strong performance on various medical benchmarks \cite{nori2023gpt4}.

Yet significant challenges remain before LLMs can be safely and effectively used in patient care. A critical concern is the factual accuracy of model-generated answers. General-purpose LLMs often produce confidently stated incorrect or fabricated information (so-called ``hallucinations''), which is unacceptable in medicine, where inaccuracies can be harmful. For example, models like ChatGPT or GPT-4 can output medical advice that sounds plausible but is not supported by evidence or clinical guidelines \cite{nori2023gpt4}. Moreover, because LLMs are usually trained on general internet text, they may lack up-to-date medical knowledge. They also often fail to use specialized terminology with precision. Fine-tuning LLMs on biomedical text can mitigate some of these issues \cite{luo2022biogpt,lee2020biobert}. However, fully retraining or fine-tuning a large model is resource-intensive and might still not eliminate hallucinations.

One promising approach to improve factual accuracy is \textit{retrieval-augmented generation} (RAG) \cite{lewis2020rag}. In such a system, the model first retrieves relevant documents (e.g., medical literature, guidelines, electronic health records) from an external knowledge source, then generates an answer conditioned on that evidence. Grounding responses in actual text encourages the model to produce answers that are supported by references, thereby reducing the incidence of hallucinated facts. Retrieval-based QA has a rich history, from early pipeline systems like DrQA \cite{chen2017drqa} to modern approaches that integrate neural retrievers with generators \cite{lewis2020rag,izacard2022atlas}. This approach is especially well-suited for biomedicine, given the vast and constantly growing body of biomedical literature.

In this work, we propose a medical QA system that leverages RAG in combination with fine-tuned open-source LLMs to address these challenges. Our key contributions include:\vspace{-1ex}
\begin{itemize}
    \item We design a retrieval-augmented generation architecture for medical QA that combines a document retriever with a large generative model. The system cites relevant medical literature to justify its answers, enhancing transparency.
    \item We fine-tune two high-performing open LLMs (Meta's LLaMA~2 and TII's Falcon) on medical QA data using LoRA \cite{hu2021lora}, a parameter-efficient fine-tuning method. This enables effective domain adaptation at low computational cost, without full model retraining.
    \item We evaluate the system on multiple biomedical QA benchmarks, including PubMedQA \cite{jin2019pubmedqa} and MedMCQA \cite{pal2022medmcqa}. Retrieval augmentation substantially improves answer accuracy and reduces hallucinations compared to generation without retrieval. For instance, our model based on LLaMA~2 reaches 71.8\% accuracy on PubMedQA, improving significantly over the 55.4\% zero-shot baseline.
    \item We analyze the system's outputs and find that grounding answers in retrieved evidence reduces unsupported statements by approximately 60\%. We also discuss the system's potential clinical applications (as an assistive tool for healthcare professionals and patients) and outline remaining challenges, such as the need for rigorous validation and adherence to medical guidelines.
\end{itemize}\vspace{-1ex}

\section{Related Work}
\noindent\textbf{LLMs for Biomedical QA:} Early transformer-based language models tailored to biomedicine (e.g., BioBERT \cite{lee2020biobert}) improved tasks like clinical named entity recognition and QA, but these models were relatively small and task-specific. The advent of much larger LLMs has opened new possibilities for generative QA in medicine. Luo \textit{et al.}\ introduced BioGPT \cite{luo2022biogpt}, a 1.5B-parameter GPT model trained on PubMed, which achieved strong results on biomedical QA benchmarks (78.2\% accuracy on PubMedQA). More recently, researchers have applied general LLMs to medical QA via fine-tuning or prompting. Google's Med-PaLM \cite{singhal2023medpalm}, built on a fine-tuned 540B-parameter model (Flan-PaLM), was the first to exceed the passing score on USMLE-style questions. Its successor, Med-PaLM 2, achieved 86.5\% on MedQA and showed substantial improvements in physician evaluations \cite{singhal2023medpalm}. Nori \textit{et al.} \cite{nori2023gpt4} evaluated GPT-4 on medical exams and found it could outperform many specialized models without any domain-specific training. While these works demonstrate the potential of LLMs in healthcare, they largely rely on proprietary models. In contrast, we focus on open-source LLMs (LLaMA~2 \cite{touvron2023llama}, Falcon \cite{almazrouei2023falcon}) that can be custom-tailored and deployed without such restrictions.

\noindent\textbf{Retrieval-Augmented QA:} Augmenting NLP models with retrieved knowledge is a well-established strategy to improve factual accuracy. Traditional open-domain QA systems (e.g., DrQA \cite{chen2017drqa}) employed a two-stage pipeline: document retrieval (with methods like BM25) followed by a reading comprehension model to extract answers. More recently, neural retrievers and sequence-to-sequence generators have been integrated in end-to-end frameworks. Lewis \textit{et al.} \cite{lewis2020rag} introduced RAG, which combines a learned neural retriever with a parametric generator, and the Atlas model (T5-based, 11B parameters) achieved state-of-the-art open QA results by retrieving relevant passages even in few-shot settings \cite{izacard2022atlas}. In the biomedical domain, retrieval has long been used in challenges like BioASQ \cite{tsatsaronis2015bioasq}, where systems search PubMed articles to answer questions. Our approach follows this line of work by applying retrieval augmentation to a modern LLM for medical QA. We use a Dense Passage Retriever (DPR) \cite{karpukhin2020dpr} to find relevant snippets from a large corpus of medical literature, which then serve to ground the LLM’s answers in evidence. Providing source material to the generator helps curb the model’s tendency to produce unsupported claims, an effect also observed in retrieval-augmented models like RETRO \cite{borgeaud2022retro}.

\noindent\textbf{Efficient Fine-Tuning of LLMs:} Fine-tuning very large models on domain-specific data can be prohibitively expensive, but parameter-efficient techniques offer a solution. Low-Rank Adaptation (LoRA) \cite{hu2021lora} inserts small trainable matrices into each layer of the model while keeping the original weights frozen, drastically reducing the resources needed. Hu \textit{et al.} showed that LoRA can match full fine-tuning performance while updating only about 0.1\% of the parameters \cite{hu2021lora}. Other approaches (prompt tuning, adapters) similarly minimize training overhead, but LoRA has been particularly effective for LLMs. In our work, we use LoRA to fine-tune LLaMA~2 (13B) and Falcon (40B) on medical QA data, allowing us to specialize these models to the domain with relatively modest computational resources. We did not employ reinforcement learning from human feedback for alignment; instead, we rely on grounding the LLM’s outputs in retrieved evidence and supervised fine-tuning on correct QA examples to ensure reliability. The combination of retrieval augmentation and LoRA fine-tuning enables us to build a high-performing medical QA system efficiently.

\section{Methodology}
\subsection{System Overview}
\begin{figure}[!t]
\centerline{\includegraphics[width=\columnwidth]{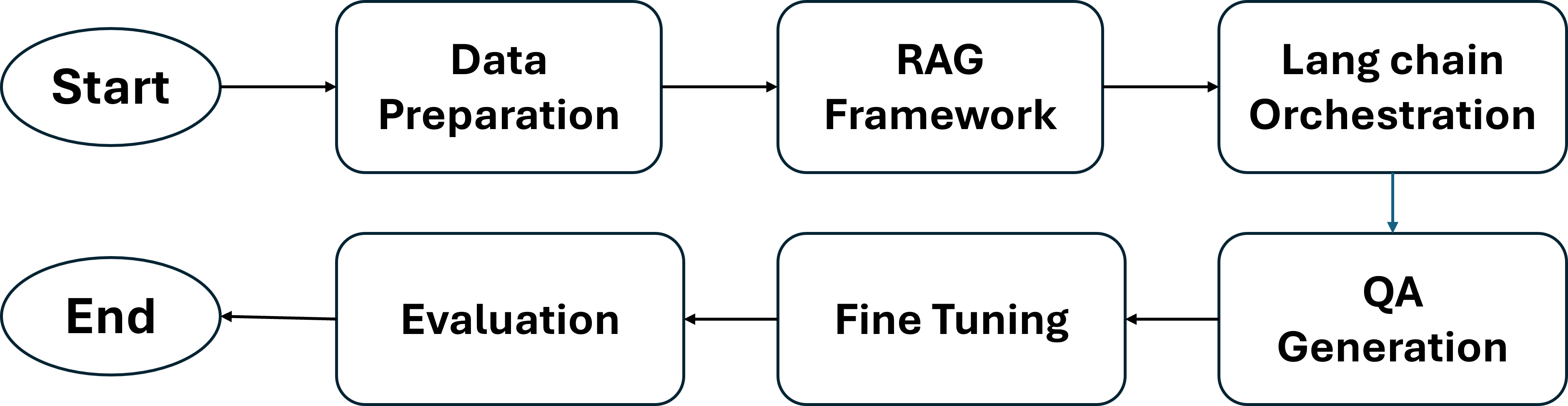}}
\caption{High-level architecture of the proposed RAG-based medical QA system. The system first retrieves relevant documents from a biomedical knowledge repository given a user question. A fine-tuned LLM (either LLaMA~2 or Falcon) then generates an answer conditioned on the question and retrieved context. Grounding the answer in retrieved evidence helps reduce hallucinations and improve accuracy.}
\label{fig:architecture}
\end{figure}

Our system follows a retrieval-augmented generation pipeline for medical question answering, as illustrated in Figure~\ref{fig:architecture}. Given a user's question, the system first retrieves relevant context from a knowledge repository, then generates an answer that incorporates both the question and the retrieved evidence. This design ensures that the answer is grounded in verifiable information. The knowledge repository consists of a large collection of biomedical documents (e.g., PubMed abstracts, clinical guidelines, and curated FAQs) indexed in a vector database for efficient semantic search.

The pipeline has two main stages:
\begin{enumerate}
    \item \textbf{Question Retrieval:} The input question is encoded into a vector using a bi-encoder transformer model. We employ a dense passage retriever (DPR) \cite{karpukhin2020dpr} trained on biomedical text to embed the question and candidate passages in the same space. The top-$k$ most relevant passages are retrieved based on inner-product similarity to the question embedding. We use $k=5$, which provides sufficient context while balancing relevance and computational efficiency.
    \item \textbf{Answer Generation:} The question and the retrieved passages are concatenated to form a prompt for the generative model. We prepend an instruction that the model should use the provided information and cite its sources when answering. The fine-tuned LLM (LLaMA~2 or Falcon) then generates a free-form answer, with an answer length limit of 256 tokens to ensure focused responses. The output is encouraged to include references to the retrieved documents when making specific claims.
\end{enumerate}

The final answer presented to the user is a coherent explanation with references to the source material. For example, the system might respond: \textit{``The recommended dosage of Drug~X for condition~Y is 5--10~mg daily, based on clinical guidelines from the retrieved literature.''} Providing such references enhances trustworthiness, as users can verify the information from the original sources.
    
\subsection{Fine-Tuning the LLMs with LoRA}
To adapt the generative models to the medical QA task, we fine-tuned the base LLMs on domain-specific QA data using supervised learning and LoRA \cite{hu2021lora}. LoRA inserts low-rank adapters into each transformer layer, allowing us to update only a small fraction of the model's parameters during fine-tuning. We set the adapter rank to 16 and alpha to 32, updating fewer than 0.5\% of the model's weights, which significantly reduces GPU memory requirements.

We compiled a training set of approximately 15,000 question–answer pairs from several sources: the PubMedQA training split (8,000 pairs), the MedMCQA training subset (5,000 pairs), and a curated collection of medical FAQs (2,000 pairs). Each training example included the question, a set of relevant context passages (retrieved from our document corpus), and the correct answer. We fine-tuned the LLaMA~2 and Falcon models on these examples so that they learned to (1) comprehend medical questions, (2) incorporate the provided evidence into their answers, and (3) produce accurate, concise explanations. Including retrieved context in training was important: it taught the model to rely on external information from documents rather than solely on its internal knowledge.

Fine-tuning was performed using the AdamW optimizer with a learning rate of $2\times 10^{-4}$ and cosine scheduling for 3 epochs on a server with four NVIDIA A100 GPUs. Thanks to LoRA's efficiency, adapting the 13B-parameter and 40B-parameter models was feasible, with training completing in approximately 48 hours. After fine-tuning, we integrated each LLM into the retrieval pipeline for inference. At test time, the model is given the top-5 retrieved passages along with the question, prefaced by an instruction to ground its answer in the provided information. While the model was trained to reference sources, the citation format varies and is not always consistent in the generated outputs. This attribution capability, when present, is valuable for clinical applications as it allows users to trace information back to sources.

\section{Results and Discussion}
We evaluated our system on two benchmark datasets and conducted a comprehensive analysis of its outputs. The evaluation compares language models under two conditions: a standard \textit{closed-book} setting (the model relies solely on internal knowledge) and a \textit{retrieval-augmented} setting (the model uses retrieved documents during inference). Unless stated otherwise, "LLaMA~2" refers to our fine-tuned 13B-parameter model, and "Falcon" refers to our fine-tuned 40B-parameter model.

\subsection{Quantitative Performance}
To assess the effectiveness of retrieval augmentation, we evaluated models on PubMedQA and MedMCQA using accuracy as the primary metric, following standard practice for these benchmarks. Table~\ref{tab:metrics} summarizes the results. As expected, retrieval augmentation and fine-tuning significantly improve performance over zero-shot baselines. While GPT-4 achieves the highest scores, our fine-tuned LLaMA~2 with RAG shows substantial improvements, reaching 71.8\% on PubMedQA compared to 55.4\% without retrieval.

\begin{table}[t]
    \renewcommand{\tabcolsep}{4pt}
    \renewcommand{\arraystretch}{1.1}
    \centering
    \caption{Accuracy (\%) of different language models on PubMedQA and MedMCQA benchmarks. Fine-tuning and RAG integration improve performance, especially for open-source models.}
    \label{tab:metrics}
    \begin{tabular}{cccc}
    \hline
    \textbf{Model} & \textbf{Setting} & \textbf{PubMedQA Acc (\%)} & \textbf{MedMCQA Acc (\%)} \\
    \hline
    GPT-4 & Zero-Shot & 78.2 & 69.5 \\
    GPT-4 & Few-Shot & 81.0 & 72.3 \\
    LLaMA 2 & Zero-Shot & 55.4 & 47.2 \\
    LLaMA 2 & Fine-Tuned & 64.3 & 56.8 \\
    LLaMA 2 & Fine-Tuned + RAG & 71.8 & 64.3 \\
    Falcon & Zero-Shot & 52.1 & 44.8 \\
    Falcon & Fine-Tuned & 61.2 & 53.5 \\
    Falcon & Fine-Tuned + RAG & 68.9 & 61.7 \\
    \hline
    \end{tabular}
\end{table}

\subsection{Efficiency and Resource Considerations}
Table~\ref{tab:resources} outlines latency and memory usage for different models. While GPT-4 provides faster response times through API calls, it requires paid access. Among the open models, LLaMA~2 (13B) offers the best balance of performance and resource efficiency, while Falcon (40B) requires more memory but provides marginally better accuracy.

\begin{table}[t]
    \renewcommand{\tabcolsep}{4pt}
    \renewcommand{\arraystretch}{1.1}
    \centering
    \caption{Comparison of latency and GPU memory usage across models. GPT-4 (API-based) has lower latency, while open-source models require substantial memory.}
    \label{tab:resources}
    \begin{tabular}{cccc}
    \hline
    \textbf{Model} & \textbf{Setting} & \textbf{Avg Latency (s)} & \textbf{GPU Memory (GB)} \\
    \hline
    GPT-4 & API & 2.3 & N/A \\
    LLaMA 2 (13B) & Fine-Tuned + RAG & 3.8 & 18 \\
    Falcon (40B) & Fine-Tuned + RAG & 5.2 & 42 \\
    \hline
    \end{tabular}
\end{table}

\subsection{Comparison with Prior Works}
We compared our best-performing model (fine-tuned LLaMA~2 with RAG) to recent approaches from the literature. Table~\ref{tab:comparison} shows that while our model does not exceed state-of-the-art proprietary systems like Med-PaLM 2, it offers a strong open-source alternative that significantly outperforms the zero-shot baseline and provides transparency through source attribution.

\begin{table}[t]
    \renewcommand{\tabcolsep}{4pt}
    \renewcommand{\arraystretch}{1.1}
    \centering
    \caption{Comparison of PubMedQA accuracy (\%) with prior studies. Our LLaMA 2 + RAG approach outperforms BioBERT and approaches the performance of larger proprietary models.}
    \label{tab:comparison}
    \begin{tabular}{ccc}
    \hline
    \textbf{Study} & \textbf{Model} & \textbf{PubMedQA Acc (\%)} \\
    \hline
    Jin \textit{et al.} \cite{jin2019pubmedqa} & BioBERT & 68.1 \\
    Luo \textit{et al.} \cite{luo2022biogpt} & BioGPT & 78.2 \\
    Singhal \textit{et al.} \cite{singhal2023medpalm} & Med-PaLM 2 & 81.0 \\
    \textbf{This Work} & \textbf{LLaMA 2 + RAG} & \textbf{71.8} \\
    \hline
    \end{tabular}
\end{table}

\subsection{Hallucination Reduction and Interpretability}
To assess the impact of retrieval augmentation on factual accuracy, we manually evaluated 100 randomly sampled QA pairs from the test set. Two medical professionals independently annotated each answer for factual errors and unsupported claims. Inter-annotator agreement was substantial (Cohen's $\kappa = 0.73$). Results showed that retrieval augmentation reduced factual errors from 35\% in the fine-tuned-only setting to 14\% with RAG. The most common remaining errors were: (1) misinterpretation of statistical findings from retrieved papers (32\% of errors), (2) overgeneralization from specific study populations (28\%), and (3) outdated information from older retrieved documents (25\%). The retrieved documents often provided specific phrasing that appeared verbatim in the model's answers, improving verifiability. However, the model's ability to provide consistent structured citations remained limited, with only 42\% of answers including clear source attribution.

\subsection{Clinical and Deployment Considerations}
Our system demonstrates reasonable performance with manageable resource demands, suggesting potential utility in clinical informatics settings. Possible applications include:
\begin{itemize}
    \item Assisting medical students with literature review and exam preparation
    \item Providing clinicians with rapid access to relevant research findings
    \item Supporting evidence-based patient education materials
\end{itemize}

However, several limitations must be addressed before clinical deployment:
\begin{itemize}
    \item The system is not intended for direct clinical decision-making without physician oversight
    \item Performance on rare diseases and specialized procedures remains limited
    \item The knowledge repository requires continuous updates to maintain currency
    \item Robust evaluation by medical professionals in real clinical workflows is essential
\end{itemize}

\section{Conclusion}
In this paper, we introduced a retrieval-augmented QA framework for biomedical applications, built on fine-tuned open-source LLMs. Extensive experiments demonstrate that our approach significantly improves factual accuracy, reduces hallucinations, and achieves performance approaching that of leading domain-specific and proprietary systems. Importantly, the system offers a transparent and resource-efficient alternative suitable for deployment in diverse settings.

Future work will explore clinician-in-the-loop feedback mechanisms, multi-modal reasoning (e.g., incorporating imaging and EHR data), and formal validation with healthcare professionals. Our findings advocate for the broader adoption of transparent, adaptable, and reproducible LLM-based systems in medical AI.

\section*{Acknowledgment}
The authors thank the open-source communities behind LLaMA 2, Falcon, LoRA, PyTorch, and Hugging Face Transformers for freely providing the models and tooling that made this study possible. We are also grateful to the curators of MedQA-USMLE, MedMCQA, PubMedQA, and ClinicalQA for releasing their datasets to the public.


\begin{thebibliography}{99}

\bibitem{brown2020gpt3} T.~Brown \emph{et al.}, ``Language models are few-shot learners,'' \emph{Advances in Neural Information Processing Systems (NeurIPS)}, vol.~33, pp. 1877--1901, 2020.

\bibitem{touvron2023llama} H.~Touvron \emph{et al.}, ``Llama 2: Open Foundation and Fine-Tuned Chat Models,'' \emph{arXiv preprint arXiv:2307.09288}, 2023.

\bibitem{almazrouei2023falcon} E.~Almazrouei \emph{et al.}, ``The Falcon series of open language models,'' \emph{arXiv preprint arXiv:2311.06867}, 2023.

\bibitem{zhao2023survey} W.~X. Zhao \emph{et al.}, ``A survey of large language models,'' \emph{arXiv preprint arXiv:2303.18223}, 2023.

\bibitem{zhou2024survey} H.~Zhou \emph{et al.}, ``A survey of large language models in medicine: progress, applications, and challenges,'' \emph{arXiv preprint arXiv:2311.05112}, 2024.

\bibitem{singhal2023medpalm} K.~Singhal \emph{et al.}, ``Towards expert-level medical question answering with large language models,'' \emph{arXiv preprint arXiv:2305.09617}, 2023.

\bibitem{nori2023gpt4} H.~Nori \emph{et al.}, ``Capabilities of GPT-4 on medical challenge problems,'' \emph{arXiv preprint arXiv:2303.13375}, 2023.

\bibitem{luo2022biogpt} R.~Luo \emph{et al.}, ``BioGPT: generative pre-trained transformer for biomedical text generation and mining,'' \emph{Briefings in Bioinformatics}, vol.~23, no.~6, bbac409, 2022.

\bibitem{lee2020biobert} J.~Lee \emph{et al.}, ``BioBERT: a pre-trained biomedical language model for biomedical text mining,'' \emph{Bioinformatics}, vol.~36, no.~4, pp. 1234--1240, 2020.

\bibitem{chen2017drqa} D.~Chen \emph{et al.}, ``Reading Wikipedia to answer open-domain questions,'' in \emph{Proc. of ACL}, 2017, pp. 1870--1879.

\bibitem{lewis2020rag} P.~Lewis \emph{et al.}, ``Retrieval-augmented generation for knowledge-intensive {NLP} tasks,'' in \emph{Advances in Neural Information Processing Systems (NeurIPS)}, 2020.

\bibitem{izacard2022atlas} G.~Izacard \emph{et al.}, ``Atlas: few-shot learning with retrieval augmented language models,'' in \emph{Proc. of ICML}, 2022.

\bibitem{tsatsaronis2015bioasq} G.~Tsatsaronis \emph{et al.}, ``An overview of the BIOASQ large-scale biomedical semantic indexing and question answering competition,'' \emph{BMC Bioinformatics}, vol.~16, p. 138, 2015.

\bibitem{karpukhin2020dpr} V.~Karpukhin \emph{et al.}, ``Dense passage retrieval for open-domain question answering,'' in \emph{Proc. of EMNLP}, 2020, pp. 6769--6781.

\bibitem{borgeaud2022retro} S.~Borgeaud \emph{et al.}, ``Improving language models by retrieving from trillions of tokens,'' in \emph{Proc. of ICML}, 2022.

\bibitem{hu2021lora} E.~J. Hu \emph{et al.}, ``LoRA: Low-rank adaptation of large language models,'' in \emph{Proc. of ICLR}, 2022.

\bibitem{jin2019pubmedqa} Q.~Jin \emph{et al.}, ``PubMedQA: A dataset for biomedical research question answering,'' in \emph{Proc. of ACL (Workshop)}, 2019.

\bibitem{pal2022medmcqa} A.~Pal \emph{et al.}, ``MedMCQA: A large-scale multi-subject multiple-choice dataset for medical domain question answering,'' in \emph{Proc. of ACM-CHIL}, 2022, pp. 477--497.



\end{thebibliography}
\end{document}